\documentclass[journal]{IEEEtran}

 \usepackage[pdftex]{graphicx}
 \usepackage{amsmath}
 \usepackage{xcolor}

\usepackage{amssymb}
\usepackage{latexsym}
\usepackage{multirow}

\usepackage{url}

\begin{document}
%
\title{A Saliency-based Convolutional Neural Network for Table and Chart Detection in Digitized Documents}
%
%
%
\author{I.~Kavasidis, S.~Palazzo, C.~Spampinato, C.~Pino, D.~Giordano, D.~Giuffrida, P.~Messina
\thanks{I. Kavasidis, S. Palazzo, C. Spampinato, C. Pino and D. Giordano are with the Pattern Recognition and Computer Vision (PeRCeiVe) Lab at the University of Catania, in Italy, email: (see www.perceivelab.com). D. Giuffrida and P. Messina are with Tab2Ex company, San Jose', California (USA), see http://www.tab2ex.com/ and https://snapchart.co.}
\thanks{Manuscript submitted April 13, 2018}}

%
%

\markboth{Submitted to IEEE Transactions on Multimedia}%
{Shell \MakeLowercase{\textit{et al.}}: Bare Demo of IEEEtran.cls for IEEE Journals}
%



\maketitle

\begin{abstract}
Deep Convolutional Neural Networks (DCNNs) have recently been applied successfully to a variety of vision and multimedia tasks, thus driving development of novel solutions in several application domains. 
Document analysis is a particularly promising area for DCNNs: indeed, the number of available digital documents has reached unprecedented levels, and humans are no longer able to discover and retrieve all the information contained in these documents without the help of automation. Under this scenario, DCNNs offers a viable solution to automate the information extraction process from digital documents. 
Within the realm of information extraction from documents, detection of tables and charts is particularly needed as they contain a visual summary of the most valuable information contained in a document. For a complete automation of visual information extraction process from tables and charts, it is necessary to develop techniques that localize them and identify precisely their boundaries.\\
In this paper we aim at solving the table/chart detection task through an approach that combines deep convolutional neural networks, graphical models and saliency concepts. In particular,  we propose a saliency-based fully-convolutional neural network performing multi-scale reasoning on visual cues followed by a fully-connected conditional random field (CRF) for localizing tables and charts in digital/digitized documents.\\
Performance analysis carried out on an extended version of ICDAR 2013 (with annotated charts as well as tables) shows that our approach yields promising results, outperforming existing models.
\end{abstract}


%
\IEEEpeerreviewmaketitle

\section{Introduction}

Production and storage of digital documents have increased exponentially in the last two decades. Extracting and retrieving information from this massive amount of data have become inaccessible to human operators and a large amount of information captured in digital documents may go lost or never seen. As a consequence, a large body of research has focused on automated methods for document analysis. Most of these efforts are directed towards the development of Natural Language Processing (NLP) methods, that analyze both grammar and semantics of text with the goal of automatically extracting, understanding and, eventually, summarizing key information from digital documents. However, while text is, inarguably, a fundamental way to convey information, there are contexts where graphical elements are much more powerful. For example, in scientific papers, many experiments, variables and numbers need to be reported in a concise way that fits better with tables/figures than text. The idiom \textit{``a picture is worth a thousand words''} describes exhaustively the power that graphical elements possess in conveying information that would be otherwise cumbersome, both for the writer to express and the reader to understand. Thus, it is of primary importance for an effective automatic document processing approach to  gather information from tables and charts. Several commercial software products that convert digitized and digital documents into processable text already exist. However, most of them either largely fail when dealing with graphical elements or require an exact localization of such elements to work properly. For this reason, a crucial pre-processing step in automated data extraction from tables and charts is to find their exact location.

The problem of identifying objects in images traditionally falls in the object detection research area, where, nowadays, Deep Convolutional Neural Networks (DCNNs) play the leading role~\cite{Redmon_2017_CVPR,7792742}. However, naively employing DCNN-based object detectors in digital documents, suitably transformed into images, leads to failures mainly because of the intrinsic appearance difference between digital documents and natural images (the data for which models are mainly thought for). Trying to train models from scratch may be unfeasible due to the large number of images required and to the lack of suitably annotated document datasets. Moreover, such approaches generally exploit the visual differences between object categories: while the visual characteristics of certain graphical elements (e.g., charts) significantly differ from text, the same cannot be said for tables, whose main differences from the surrounding content lie mostly in the layout. Finally, many of the existing object detectors are often prone to potential errors by upstream region proposal models and are not able to detect simultaneously all the objects of interest in an image \cite{ren2015faster,Girshick_2015_ICCV}.

For all these reasons, traditional object detectors may be inappropriate to perform accurate table and chart detection/segmentation. Given our previous consideration on how tables are identifiable mostly by their layout than by their content (which is still textual), a possible solution would be that of posing the problem as a \emph{saliency detection} one. Saliency is the perceptual quality of certain objects (or object parts) that possess distinctive features with respect to the surroundings: such salient characteristics may not be very different from other elements in the scene, but may attract the viewer's attention through aspects such as, for example, arrangement~\cite{Itti1998}. This definition seems particularly appropriate to tables, whose organized structure easily stands out in a page otherwise filled with an unstructured flow of text. We argue that identifying local (i.e., pixelwise) patterns such as lines and spacings that characterize the presence of a table is the key to localizing the whole object. In our case, however, we focus on different objects (not only tables, but also charts) and thus we have different types of saliency, corresponding to the targeted object categories: from this point of view, the problem becomes one of \emph{semantic segmentation}, i.e., assigning a class to each pixel in an image.

DCNN-based object detection methods have been repurposed for semantic segmentation by leveraging the learned low and middle level region representations. However, the multi-scale aggregation and down-sampling employed by DCNNs for object detection show several drawbacks when performing dense prediction, which, instead, requires multi-scale reasoning at high image resolution~\cite{YuKoltun2016}. In other words, the responses at the final layers of CNN-based object detection methods are not localized enough for pixelwise classification because their invariance properties make them more suitable for high level tasks. This limitation is particularly problematic for table detection, where long-range dependencies are essential to distinguish between, for example, two horizontally-stacked tables or two groups of columns of the same table (and analogously in the vertical direction).

In this paper we propose a fully-convolutional neural network for table and chart detection able to overcome the limitations of the existing approaches. In detail, we frame our approach as a semantic image segmentation one, i.e., we perform pixelwise dense prediction assigning to each pixel a likelihood of being part of an object of interest. In order to capture long-term dependencies between elements in a document, our network exploits dilated convolutions \cite{YuKoltun2016} for effectively extracting and using multi-scale contextual information in a dense prediction problem. Since tables and charts are among the most salient (according to visual saliency definition) areas in an image, our network is first pre-trained on saliency detection datasets to learn basic visual cues and afterwards fine-tuned on digital documents. Additionally, in order to provide a stronger supervision to the internal saliency detectors, we employ the approach introduced in~\cite{MURABITO2018}, by adding a loss term related to the capability of the saliency maps to identify regions that are distinctive for visual classification in one of the four target categories. Finally, predictions of our network are enhanced with a fully connected Conditional Random Field (CRF) ~\cite{NIPS2011_4296}.

Performance evaluation, carried out on an extended version of ICDAR 2013 dataset (with annotated charts as well as tables), reveals that our network significantly outperforms other DCNN models as well as state-of-the-art methods in detecting tables and charts in digital documents.

The paper is organized as follows: in the next section a review of the literature on table detection/classification is presented, while in Section III our saliency-based semantic segmentation approach is described. In Section IV, we report the results yielded by our approach on the extended ICDAR 2013 dataset, and in the final section conclusions are drawn and future directions given.

\section{Related Work}
Given the large quantity of digital documents that are available today, it is mandatory to develop automatic approaches to extract, index and process information for long-term storage and availability. Consequently, there is a large body of research on document analysis methods attempting to extract different types of objects (e.g., tables, charts, pie charts, etc.) from various document types (text documents, source files, documents converted into images, etc.).

Before the advent of deep learning, most works on document analysis for table detection were based on exploiting \textit{a priori} knowledge on object properties by analyzing tokens extracted from source document files \cite{chen2011table,fang2011table,shafait2010table,gatos2005automatic,deivalakshmi2014detection}. For example, \cite{fang2011table} proposes a method for table detection in PDF documents, which uses tags of  tabular separators to identify the table region. Similarly, in \cite{shafait2010table}, tables are detected by thoroughly analyzing the page layout and searching for tab-stop tokens (i.e., vertical lines that delimit areas of text), which are then combined to see if they match a set of predefined criteria related to tabular shape and, accordingly, to decide whether a candidate area is a table or not.
Of course, the main shortcoming of all methods that rely on detecting horizontal or vertical lines for table detection is that they fail to identify tables without borders.
Alternatively, methods operating on image conversion of document files and exploiting only visual-cues for table detection have been proposed \cite{krupl2005using,krupl2006visually, jahan2014locating,liu2008identifying,raskovic2012borderless}.
For example, \cite{krupl2006visually} describes an approach for extracting tables from web documents without searching for HTML \textit{table} tags, instead using visual cues and heuristic rules. However, methods of this kind have shown rather limited performance for specific table categories~\cite{jahan2014locating,liu2008identifying,raskovic2012borderless} and are not general enough for consumer market.
Similar computer vision--based methods have been proposed for detecting other types of graphical elements (e.g., charts, diagrams, etc.) than tables \cite{huang2003model,shao2005recognition, huang2007extraction}. These methods basically employ simple computer vision  techniques (e.g., connected components, fixed set of geometric constraints, edge detection, etc.) to extract chart images, but, as for the table detection case, they show scarce generalization capabilities. 
Low-level visual cues (e.g., intensity, contrast, homogeneity, etc.) in combination with shallow machine learning techniques have been used for specific object classification tasks \cite{karthikeyani2012machine,perez2016tao} with fair performance, but these methods are mainly for classification as they tend to aggregate global features in compact representations which are less suitable for performing object detection. 
Performance improvement has been also sought by resorting to graphical models~\cite{shetty2007segmentation,delaye2012text,pinto2003table}) such as Conditional Random Fields (CRFs), but despite their capabilities to capture fine edge details, these methods are still not as effective as expected. Our hypothesis is that the main reason for unsatisfactory performance is that tables (mainly) and charts usually cover large areas and, as such, they need methods able to account for long-range dependencies.

%
%
in \cite{pinto2003table}, Conditional Random Fields are used to label table rows in government style documents and extract the data represented achieving almost perfect performance ($>95\%$), but the main drawbacks are the simplicity of the layout and the strict adherence to a specific document template. 

With the recent rediscovery of deep learning, in particular convolutional neural networks, and its superior representation capabilities for high-level vision tasks, the document analysis research community started to employ DCNNs for document processing, with a particular focus on document classification \cite{hao2016table,afzal2015deepdocclassifier,kang2014convolutional,roy2016generalized} or object (mainly chart) classification -- after accurate manual detection \cite{liu2015chart,tang2016deepchart}. 
One recent work presenting a DCNN exclusively targeted to table detection is \cite{gilani2017table}, which employs Faster R-CNN~\cite{ren2015faster} for object detection. Nevertheless, this method suffers from the limitations mentioned in the previous section, i.e., its performance is negatively affected by the region proposal mistakes and it does not provide multiple detections for each image~\cite{tran2015table}.
In this paper, we tackle the table and chart detection problem from a different perspective, i.e., we pose it as semantic image segmentation problem, by densely predicting --- according to visual saliency principles --- for each pixel of the input image the likelihood of being part of a set of predefined classes. Despite contributing to the semantic image segmentation and saliency detection fields is out of the scope of this paper, we review briefly the state of the art on the two topics to provide the reader a means to understand the CNN architectural choices discussed in the next section. 

Saliency detection is a long researched topic and it aims at reproducing the human early unconscious process for scanning a visual scene through an attention mechanism guided by some coarse visual stimuli, followed by a late top-down process biasing the observation towards those regions that consciously attract human attention according to a specific task. The literature on this topic is large~\cite{li2016visual,kummerer2014deep,jiang2015salicon,he2015supercnn,liu2016dhsnet,li2015efficient,zhao2015saliency,liu2016dhsnet},  and the current state of the art \cite{pan2016shallow,jiang2015salicon} is oriented to fully-convolutional CNN architectures processing images at different scales for dense saliency prediction of each pixel. In addition, driving fully-convolutional CNN saliency detectors with specific goals results in improved detection performance~\cite{Almahairi2016,MURABITO2018}.

Semantic image segmentation is, instead, the task of assigning a label from a set of classes to each pixel of the image. It has been widely investigated in past years with methods aiming at finding a graph structure over the image to capture the context of each pixel by using Markov Random Fields (MRF) or Conditional Random Fields (CRF) \cite{6247739}. However, these methods employ hand-crafted features for classification, thus do not generalize well. Recently, the problem has been tackled by proposing fully-convolutional networks obtained by transforming fully-connected layers of pre-trained CNNs into convolutional layers. These methods, though more effective than traditional approaches, suffer from the scarce representation capabilities of pre-trained CNN-based classifiers for dense prediction. Indeed, they tend to create compact embeddings, via pooling and subsampling layers that aggregate multi-scale contextual information for global prediction, while dense prediction requires an explicit link between context information and pixel-wise prediction.

In this paper, we borrow methods from both research topics and propose a DCNN-based fully-convolutional approach, which exploits their respective potentialities improving, at the same time, their limitations. In detail, our approach employs a fully-convolutional architecture designed to perform semantic segmentation in a way that, through dilated convolutions \cite{YuKoltun2016}, makes an explicit assumption on the link between context information and dense prediction. This architecture is adapted to compute heatmaps containing pixelwise scores of how salient each pixels is with respect to $K$ possible object classes. In order to drive the model to highlight not only salient regions (which may be distinctive but sparse) but also the whole area of a target object, we jointly train a set of binary classifiers that employ the computed saliency maps to correctly classify regions of the input images, thus causing the classification loss to be propagated to the saliency detectors as an additional error signal~\cite{MURABITO2018}. Finally, we enhance the output maps of object location with the fully connected CRF proposed in~\cite{NIPS2011_4296}.
\section{Deep Learning Models for Table and Chart Detection and Classification}
\begin{figure*}[ht]
	\begin{center}
		\includegraphics[width=7in]{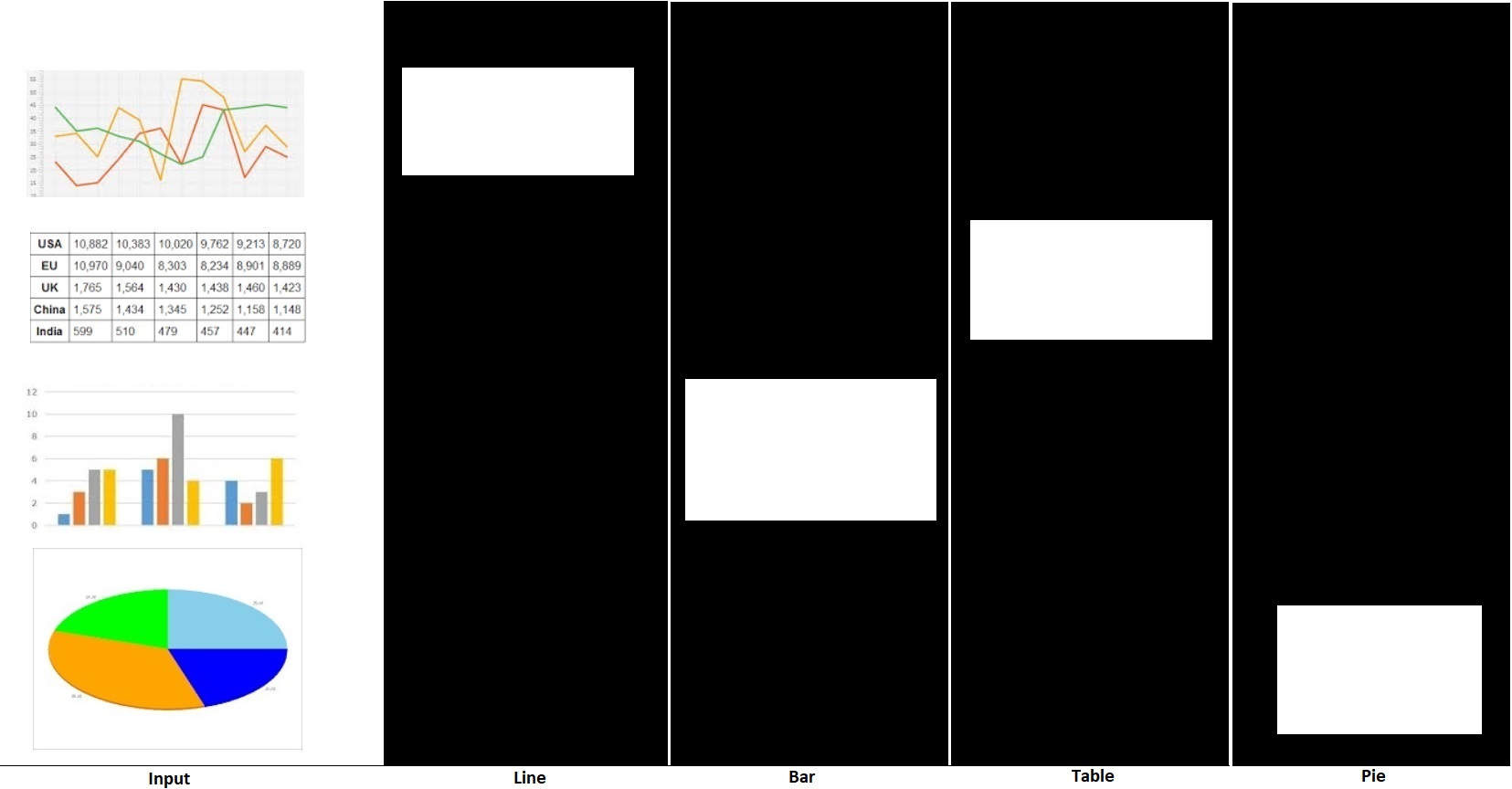}
		\caption{An example input image (first image, left) and the corresponding desired outputs (binary images, right).}
		\label{example_detection}
	\end{center}
\end{figure*}

\begin{figure*}[ht]
	\begin{center}
		\includegraphics[width=1\textwidth]{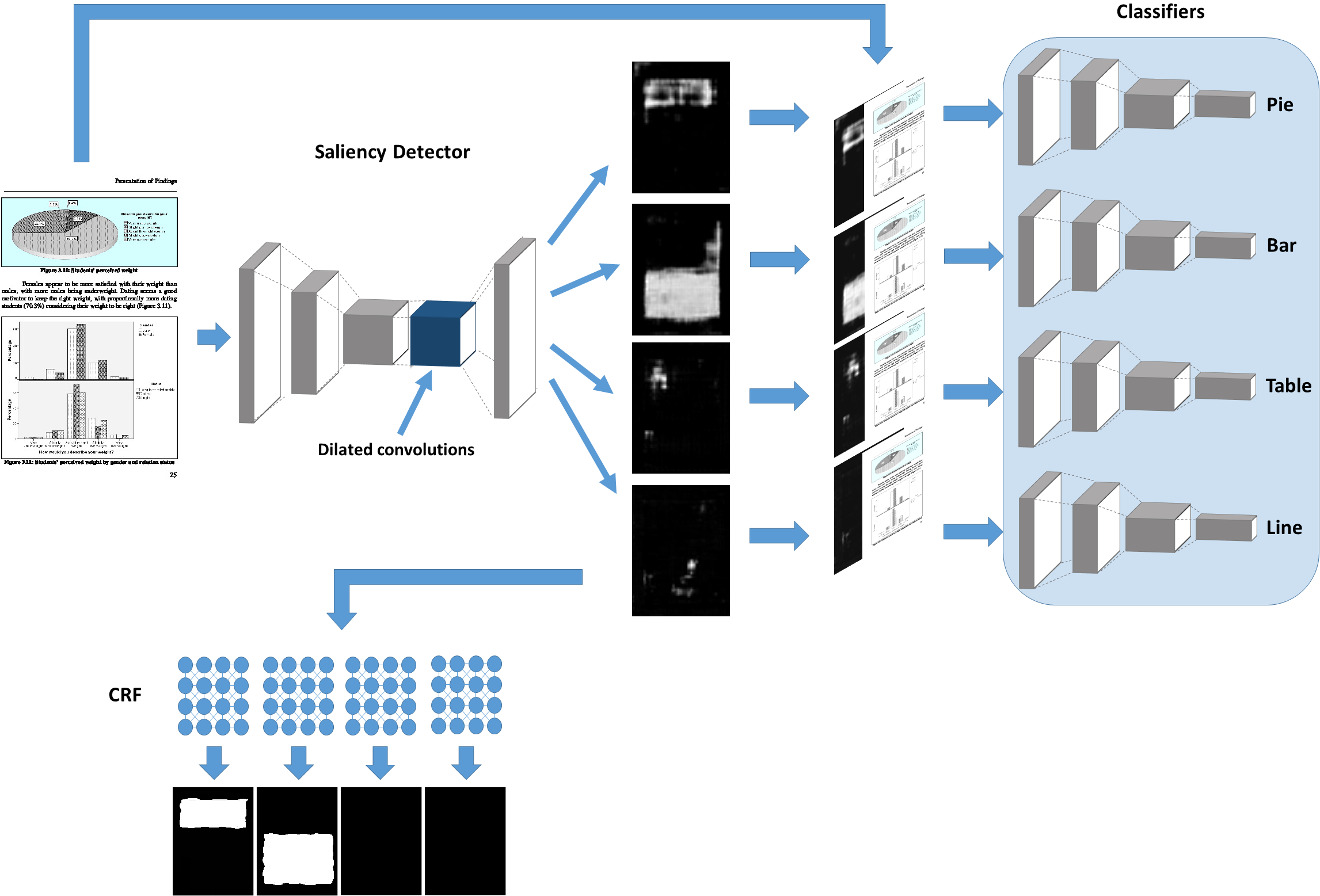}
		\caption{The proposed system. An input document is fed to a convolutional neural network trained to extract class-specific saliency maps, which are then enhanced and smoothed by CRF models. Moreover, binary classifiers are trained to provide an additional loss signal to the saliency detector, based on how useful the computed saliency maps are for classification purposes.}
			
		\label{autoencoder}
	\end{center}
\end{figure*}

The approach for table and chart extraction presented here works by receiving an image as input and generating a set of binary masks as output, from which bounding boxes are drawn. Each binary mask corresponds to the pixels that belong to objects of four specific classes: \textit{tables, pie charts, line charts} and \textit{bar charts}. An example of the expected outputs is shown in Fig.~\ref{example_detection}. The main processing engine driving our approach consists of a fully-convolutional deep learning model that performs table/chart detection and classification, followed by a conditional random field for enhancing and smoothing the binary masks (see Fig.~\ref{autoencoder}). 

\subsection{Table and chart extraction and classification} 

Given an input document page transformed into an RGB image and resized to 300$\times$300, the output of the system consists of four binary masks, one for each of the aforementioned classes. Pixels set to 1 in a binary mask identify document regions belonging to instances of the corresponding class, while 0 values are background regions (e.g., regular text).

We leverage visual saliency prediction to solve our object detection problem. As a result, the first processing block in our method is a fully-convolutional neural network (i.e., composed only by convolutional layers~\cite{long2015fully}) that extracts four class-specific heatmaps from document images.

Our saliency detection network is based on the feature extraction layers of the VGG-16~\cite{Simonyan14c} architecture. However, we applied a modification aimed at exploiting inherent properties of tables and charts: in particular, the first two convolutional layers do not employ traditional square convolution kernels, as in the original VGG-16 implementation, but use rectangular ones instead, of sizes 3$\times$7 and 7$\times$3 (equally distributed in the number of feature maps for each layer, see Table~\ref{tab_saliency_detector}). 
This set up gives our network the ability to extract table-related features (e.g., lines, spacings, columns and rows) even at the early stages. Padding was suitably added in order to keep the size of the output feature maps independent of the size of the kernels. 

\begin{table}[!th]
	\begin{center}
		\caption{Architectural details of the saliency detection network. All convolutional layers except the last one are followed by batch normalization layers and ReLU activations.}
		\begin{tabular}{|c|c|c|c|c|c|c|}
			\hline
			Layer &  Maps & Kernel size & Stride & Pad & Dilation \\
			\hline\hline
			Conv & 64 & 3$\times$7, 7$\times$3 & 1$\times$1 & 1$\times$3, 3$\times$1 & 1$\times$1 \\
			\hline 
			Conv & 64 & 3$\times$7, 7$\times$3 & 1$\times$1 & 1$\times$3, 3$\times$1 & 1$\times$1 \\
			\hline
			MaxPool & - & 2$\times$2 & 2$\times$2 & - & -\\
			\hline
			Conv & 128 & 3$\times$3 & 1$\times$1 & 1$\times$1 & 1$\times$1 \\
			\hline
			Conv & 128 & 3$\times$3 & 1$\times$1 & 1$\times$1 & 1$\times$1 \\
			\hline
			MaxPool & - & 2$\times$2 & 2$\times$2 &- & -\\
			\hline
			Conv & 256 & 3$\times$3 & 1$\times$1 & 1$\times$1 & 1$\times$1 \\
			\hline
			Conv & 256 & 3$\times$3 & 1$\times$1 & 1$\times$1 & 1$\times$1 \\
			\hline 
			Conv & 256 & 3$\times$3 & 1$\times$1 & 1$\times$1 & 1$\times$1 \\
			\hline			
						
			Conv & 512 & 3$\times$3 & 1$\times$1 & 1$\times$1 & 1$\times$1 \\
			\hline
			Conv & 512 & 3$\times$3 & 1$\times$1 & 1$\times$1 & 1$\times$1 \\
			\hline
			Conv & 512 & 3$\times$3 & 1$\times$1 & 1$\times$1 & 1$\times$1 \\
			\hline 
			Conv & 512 & 3$\times$3 & 1$\times$1 & 1$\times$1 & 1$\times$1 \\
			\hline
			Conv & 512 & 3$\times$3 & 1$\times$1 & 1$\times$1 & 1$\times$1 \\
			\hline
			Conv & 512 & 3$\times$3 & 1$\times$1 & 1$\times$1 & 1$\times$1 \\
			\hline
			DilConv & 512 & 3$\times$3 & 1$\times$1 & 1$\times$1 & 2$\times$2 \\
			\hline
			DilConv & 512 & 3$\times$3 & 1$\times$1 & 1$\times$1 & 4$\times$4 \\
			\hline
			DilConv & 512 & 3$\times$3 & 1$\times$1 & 1$\times$1 & 8$\times$8 \\
			\hline
			DilConv & 512 & 3$\times$3 & 1$\times$1 & 1$\times$1 & 16$\times$16 \\
			\hline
			Conv & 256 & 3$\times$3 & 1$\times$1 & 1$\times$1 & 1$\times$1 \\
			\hline
			Conv & 4 & 1$\times$1 & 1$\times$1 & - & 1$\times$1 \\
			
			\hline
		\end{tabular} 
		\label{tab_saliency_detector}
	\end{center}
\end{table}

After the cascade of layers from the VGG-16 architecture, the resulting 75$\times$75 feature maps are processed by a \emph{dilation block}, consisting of a sequence of dilated convolutional layers~\cite{YuKoltun2016}. While the purpose of the previous layers is that of extracting discriminative local features, the dilation block exploits dilated convolutions to establish multi-scale and long-range relations. Dilation layers increase the receptive field of convolutional kernels while keeping the feature maps at a constant size, which is desirable for pixelwise dense prediction as we do not want to spatially compact features further. The output of the dilation block is a 4-channel feature map, where each channel is the saliency heatmap for one of the target object categories. After the final convolutional layer, we upsample the 75$\times$75 maps back to the original 300$\times$300 using bilinear interpolation.

In theory, the saliency maps could be the only expected output of the model, and we could train it by just providing the correct output as supervision. However, \cite{MURABITO2018} recently showed that posing additional constraints to saliency detection --- for example, forcing the saliency maps to identify regions that are also class-discriminative --- improves output accuracy. This is highly desirable in our case as output saliency maps may miss non-salient regions (e.g., regular text) inside salient regions (e.g., table borders), while it is preferable to obtain maps that entirely cover the objects of interest. 

For this reason, we add a classification branch to the model. This branch contains as many binary classifiers as the number of target object classes. Each binary classifier receives as input a crop of the original image around an object (connected component) in the saliency detector outputs, and aims at discriminating whether that crop contains an instance of target class.
The classifiers are based on the Inception model \cite{szegedy2015going} and are architecturally identical, except for the final classification layer which is replaced by a linear layer with one neuron followed by a sigmoid nonlinearity.

\subsection{Multi-loss training}

To train the model we employ a multi-loss function that combines the error measured on the computed saliency maps with the classification error of the binary classifiers.
The saliency loss function measured between the computed saliency maps $\textbf{Y}$ (expressed as a N$\times$300$\times$300 tensor, with N being the number of object classes, 4 in our case) and the corresponding ground-truth mask $\textbf{T}$ (same size) is given by the mean squared error between the two:
\begin{equation}
\mathcal{L}_S(\textbf{Y},\textbf{T}) = \frac{1}{N\cdot h \cdot w}\sum_{k=1}^{N}\sum_{i=1}^{h}\sum_{j=1}^{w} (Y_{kij} - T_{kij})^2
\end{equation}
where \textit{h} and \textit{w} are the image height and width (in our case, both are 300), and \textit{$Y_{kij}$} and \textit{$T_{kij}$} are the values of the respective tensors at location $(i,j)$ of the $k$-th saliency map.

The binary classifiers are first trained separately from the saliency network, so that they can be used to provide a reliable error signal to the saliency detector. Training is performed using original images cropped with ground-truth annotations. For example, for training the table classifier, we 
use table annotations (available in the ground truth) and crop input images so as to contain only tables: these are the ``positive samples''. ``Negative samples'' are, instead, obtained by cropping the original images with annotations from other classes (pie chart, bar chart and line chart) or with random background regions. This procedure is performed for each classifier to be trained.
Since cropping may result in images of different sizes, all images are resized to 299$\times$299 to fit the size required by the Inception network.

Each classifier is trained to minimize the negative log-likelihood loss function:
\begin{equation}
\mathcal{L}_{C_i}(\textbf{I}, t_i) = -t_i \log C_i(\textbf{I}) - (1 - t_i) \log \left( 1 - C_i(\textbf{I}) \right)
\end{equation}

where $C_i$ ($1 \le i \le 4$) is the classifier for the $i$\textsuperscript{th} object class, and returns the likelihood that an object of the targeted class is present in image $\textbf{I}$: $t_i$ is the target label, and is 1 if $i$ is the correct class, 0 otherwise.
After training the classifiers, they are used to compute the classification loss for the saliency detector, as follows:
\begin{equation}
\mathcal{L}_C = \sum_{i=1}^N \mathcal{L}_{C_i}(\textbf{I}, t_i)
\end{equation}

The saliency detector, in this way, is pushed to provide accurate segmentation maps so that whole object regions are passed to the downstream classifiers, Indeed, if the saliency detector is not accurate enough in identifying tables, it will provide incomplete tables to the corresponding classifier, which may be then misclassified as non-table objects with a consequent increase in loss.
Note that while training the saliency detector, the classifiers themselves are not re-trained, and are only used to compute the classification loss. This prevents the binary classifiers to learn to recognize objects from their parts, thus forcing the saliency network to keep improving its detection performance. 

The multi-loss used for training the network in an end-to-end manner is, thus, given by the sum of the terms $\mathcal{L}_{C}$ and $\mathcal{L}_{S}$:
\begin{equation}
\mathcal{L} = \mathcal{L}_{C} + \mathcal{L}_{S}
\end{equation}

\subsection{Mask enhancement by Fully-Connected Conditional Random Fields} 

The outputs of our fully convolution network, usually, show irregularities such as spatially-close objects fused in one object or one object oversegmented in multiple parts. 
Improving noisy segmentation maps has been usually tackled with conditional random fields (CRFs)~\cite{kohli2009robust}, which enforce same-label assignments to spatially close pixels. 
Recently, combining CRFs with deep convolutional networks has gained increased interest to enhance segmentation outputs~\cite{chen2016deeplab,liu2015crf}. In particular, in these methods, fully-connected CRFs~\cite{chen2016deeplab} are used to recover object structures rather than smoothing segmentation outputs, as instead performed by previous methods based on short-range CRFs.  Building on the advantages provided by fully-connected CRFs, we integrate in our system a downstream module based on the architecture proposed in~\cite{krahenbuhl2011efficient}, which has demonstrated good capabilities in recovering or significantly mitigating segmentation errors.
In detail, the fully-connected CRF module adopts the following energy function:

\begin{equation}
E(x,z) = \sum_i \theta_i (x_i,z_i) + \sum_{ij} \theta_{ij} (x_i,x_j)
\end{equation}
with $x_i, x_j \in \mathcal{X}$, which is the latent space of pixel label assignment.
As unary potential $\theta_i(x_i)$ we employ the label probability likelihood of pixel $i$ provided by the upstream CNN, i.e., the output of the saliency detection network. The pairwise potential is computed, similarly to \cite{Chen2015}, as:

\begin{equation}
\begin{split}
\theta_{ij} (x_i,x_j) = \mu(x_i,x_j) \cdot \Bigg[ w_1 \cdot e^{(- \frac{||s_i - s_j||^2}{2\sigma_{\alpha}^2} - \frac{||c_i -c_j||^2}{2\sigma_{\beta}^2})} \\ + w_2 \cdot e^{(- \frac{||s_i-s_j||^2}{2\sigma_{\gamma}^2})} \Bigg]
\end{split}
\end{equation}

$\mu(x_i,x_j)=1$ if $x_i \neq x_j$, i.e. in case the two pixels have different labels,  otherwise 0, as in the Potts model~\cite{wu1982potts}. ${s}_i$, ${c}_i$ are, respectively, the spatial coordinates and RGB color values of pixel $i$, and ${s}_j$, ${c}_j$  the same values for pixel $j$. The kernel controlled by the $w_1$ weight takes into account the visuo-spatial distances between the two pixels in order to force pixels with similar colors and spatially close to get the same label. The kernel controlled by $w_2$ depends only on the spatial distance, thus enforcing smoothness between close pixels. $\sigma_{\alpha}$, $\sigma_{\beta}$ and $\sigma_{\gamma}$ are the kernel widths. Gaussian kernel is used for two main reasons: a) it enforces smoothness in learning the function controlling the data; b) it is suitable to efficient approximate probabilistic inference in fully-connected graphs~\cite{NIPS2011_4296}.
It has to be noted that CRF-based output enhancing is performed independently for each mask computed by the saliency detector.

\begin{figure*}[!ht]
	\begin{center}
		\includegraphics[width=7in]{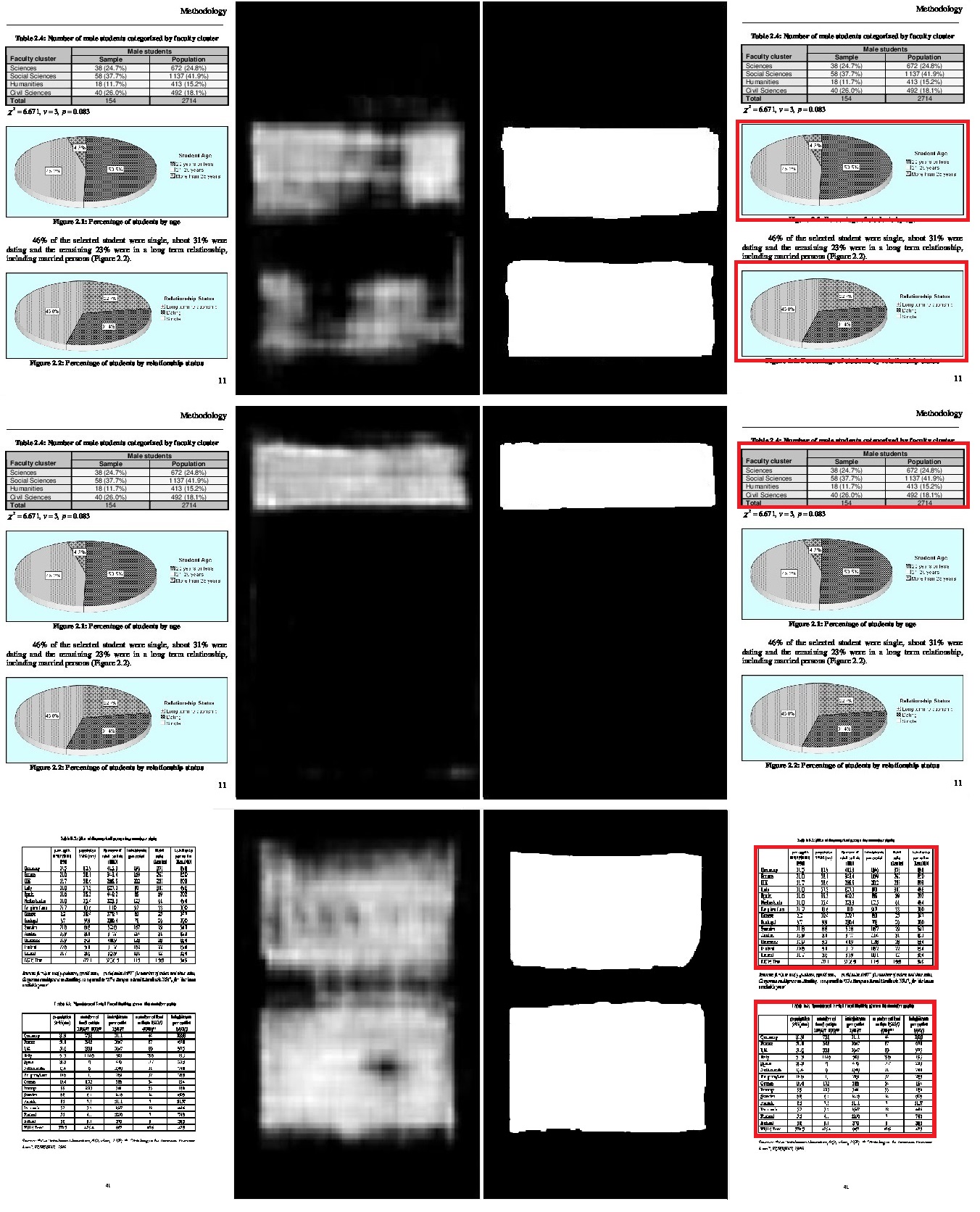}
		\caption{A detailed example output of two images taken from the ICDAR 2013 dataset. The saliency detector network extracts from the input image (\textit{left}) the class--specific saliency masks (\textit{second})(\textit{top}: pie charts, \textit{middle and bottom}: tables), which in turn are fed into a CRF model where the final segmentation masks are obtained (\textit{third}) and the bounding boxes drawn (\textit{right}). While in the middle row example the CRF module did not influence substantially the final result, it instead enhanced it in the top and bottom ones.}
		\label{example_pie}
	\end{center}
\end{figure*}

An example of input images and the corresponding intermediate and final outputs is shown in Fig.~\ref{example_pie}.
\section{Performance Analysis}
\subsection{Dataset}
\noindent \textbf{Training data collection}. To train our method, we developed a web crawler that searched and gathered images from Google Images, using the following queries: ``\emph{tables in documents}'', ``\emph{pie charts}'', ``\emph{bar charts}'', ``\emph{line charts}''. Retrieved documents were then converted to images, resulting in 50,466 images available for training. 

\noindent \textbf{Test data collection}. The dataset used for testing the models was the standard ICDAR 2013 dataset \cite{karatzas2013icdar}. However, in its original form, ICDAR 2013 does not contain annotations on other graphical elements than tables. Since our method deals also with such other types of graphical elements, we extended the ICDAR 2013 dataset by annotating also pie, line and bar charts.

\noindent \textbf{Post-processing}. Given the nature of the method employed for gathering the training dataset, we had to ensure that none of the retrieved images was already contained in the test set. To avoid duplication between training and test sets, we computed HOG features for all available images (training and test) and applied \textit{k-Nearest Neighbours}, with $k=10$, for similarity search in the two sets. This led to the removal of 24 images from the training set, for a remaining total of 50,442 training images.

\noindent \textbf{Annotations and splits}. The annotation of chart objects in training and test images was carried out by paid annotators using an adapted version of the annotation tool in \cite{kavasidis2014innovative}. 
In particular, among the 50,442 retrieved images only 19,564 had at least one object of interest, while the remaining 30,878 images did not. The 19,564 images with positive instances had in total 22,544 annotations, whose distribution is given in Table \ref{tab_icdar_extended_dataset}. Of the set of retrieved images (and related annotations), 10\% were used as a validation set for model selection, while the remaining 90\% as training set. The distribution of instances of the four target classes between the training and validation sets were approximately equal.

As for the test dataset, only 161 out of 238 images from ICDAR 2013 contained objects of interest. In these 161 images, there were 156 tables (with annotations already available) and 58 charts (of either ``pie'', ``bar'' and ``line'' types), as shown again in Table \ref{tab_icdar_extended_dataset}.

The training dataset is subject to license restrictions and cannot be published. The extended ICDAR 2013 dataset is publicly available and can be found at \textit{http://perceive.dieei.unict.it}.

\begin{table}[h!]
	\begin{center}		
		\caption{Training, Validation and Test Datasets}
		\begin{tabular}{|c|c|c|c|c|}
			\hline
			& \multicolumn{3}{|c|}{Web gathered} & ICDAR 2013\\
			\hline
			& Train & Validation & Total & Test \\		
			\hline\hline			
			Tables  & 8,780 & 976 & 9,756 & 156 \\
			\hline
			Pie charts & 1,889 & 210 & 2,099 & 9 \\
			\hline 
			Bar charts  & 5,129 & 570 & 5,699 & 29 \\
			\hline
			Line charts  & 4,491 & 499 & 4,990 & 20 \\
			\hline
			\textbf{Total annotations} & \textbf{20,289} & \textbf{2,255} & \textbf{22,544} & \textbf{214}\\
			\hline
			\hline
			\textbf{Total images}  &  \textbf{45,398} &  \textbf{5,044} & \textbf{50,442} & \textbf{238} \\
			\hline			
		\end{tabular}
		\label{tab_icdar_extended_dataset}
	\end{center}
\end{table}

\subsection{Training details, pre- and post-processing}

Since the number of annotations in the training set is not enough to avoid overfitting, we pre-trained (100 epochs) our saliency detector on the SALICON dataset \cite{jiang2015salicon}, containing saliency data of human subjects while visualizing natural images.
The binary classifiers were initialized using the weights of the Inception network pre-trained on the ImageNet dataset~\cite{deng2009imagenet}, except for the final classification layer, which is trained from scratch.
After the initial pre-training phase, the saliency detector and the classifiers were trained in an end-to-end fashion using an image as input and a) the annotation masks as training targets for the saliency detector, and b) presence/absence labels of target objects on image crops for the classifiers. The input image resolution was set to 300$\times$300 pixels. The training phase ran for 45 epochs, which in our experiments was the point where the performance of the saliency network on the validation set stopped improving.
 
All networks were trained using the Adam optimizer \cite{kingma2014adam} (learning rate was initialized to 0.001, momentum to 0.9 and batch size to 32).  

CRF training was decoupled from the saliency-based network training. In particular, after training our saliency CNN, we computed the CRF unary terms and used them for CRF training, for which we used the code in \cite{krahenbuhl2011efficient}. Parameters $w_1$, $w_2$, $\sigma_{\alpha}$, $\sigma_{\beta}$ and $\sigma_{\gamma}$ were set, respectively, to 5, 3, 50, 3, 3.

The deep-learning models were trained using three Titan X Pascal video cards. The CRF models were trained on a server with two 8-core Intel Xeon processors and 128 GB of memory.

\subsection{Performance metrics}

Our evaluation phase aimed at assessing the performance of our DCNN in localizing precisely tables and charts in digital documents, as well as in detecting and segmenting table/chart areas.

\begin{itemize}
	\item{\bf Table/chart localization performance}. To test localization performance we computed precision $Pr$, recall $Re$ and $F_1$ score by calculating true positives, false positives and false negatives. A true positive (TP) was identified when a bounding box detected by our approach overlapped (over a threshold value) a ground truth annotation. A false positive (FP) was instead defined as an object detected by our method that did not sufficiently overlap a corresponding annotation in the ground truth. A false negative (FN) was an object present in the ground truth that was not detected by our method (see Fig.~\ref{example_detection}). Then, precision, recall and $F_1$ score are calculated as follows: 
	\begin{equation} Pr = \frac{TP}{TP + FP}\end{equation}
	\begin{equation} Re = \frac{TP}{TP + FN}\end{equation} 
	\begin{equation} F_1 = \frac{2 \times Pr \times  Re}{Pr + Re}\end{equation}	

	\item {\bf Segmentation accuracy} was measured by intersection over union (IOU). While the detection scores above represent the ability of the models to detect the tables and charts that overlap with the ground truth over a certain threshold, IOU measures per-pixel performance by comparing the exact number of the pixels that were detected as belonging to a table or chart. In other words, the IOU score reflects the accuracy in finding the correct boundaries of table and chart regions. IOU is computed by comparing the segmentation mask $M_\text{O}$ provided by our model, and the corresponding ground truth mask $M_\text{GT}$ as follows:
	
	\begin{equation}
		\label{eq_iou}
		IOU_O = \frac{M_\text{O} \cap M_\text{GT} }{M_\text{O} \cup M_\text{GT}}.			
	\end{equation}

	Higher IOU score values mean that there is a substantial overlap between the detected table/chart and the annotation. The final IOU score reported in the following results is the average IOU values on all ground truth tables and charts. 

\end{itemize}

The proposed model consists of several functional blocks (saliency detection, binary classification, CRF) which are stacked together for final prediction. In order to assess how each block influenced performance, we carried out an ablation study by calculating the above metrics under the following configurations:
\begin{enumerate}
	\item \textbf{SAL}: this is the baseline configuration and refers to using only the saliency network (SAL):
	\item \textbf{SAL-CRF}: this configuration, instead, consists of the saliency network followed by the fully-connected CRF;
	\item \textbf{SAL-CL}: saliency detector followed by the binary classifiers;
	\item \textbf{ALL}: the whole system including all parts (saliency detection, CRF and binary classifiers).
\end{enumerate}

\subsection{Results}

Table detection performance was tested on the standard ICDAR 2013 dataset, while chart detection performance on the extended ICDAR 2013. 

For table/chart localization results, we set IOU = 0.5 as overlap threshold to identify true positives, following common protocols in object detection~\cite{everingham2010pascal}.

The results obtained by the four configurations previously described are reported in Table \ref{tab_performance_aal}. Our system performed very well in all object types, and this performance increased progressively from the baseline configuration (SAL) to the more complex architectures (i.e, SAL-CRF, SAL-CL and ALL). In particular, the baseline configuration (SAL) achieved an average $F_1$ score of 69.0\%, with the top performance achieved on the ``Tables'' category (76.3\%) and the worst on the ``Line charts'' category (63.4\%).
The addition of the CRF model (SAL-CRF) to the baseline architecture increased the average $F_1$ score by about 9\%. The increase came mainly through a sharper increase in the recall (i.e., reduction of false negatives) of about 12\%, especially for chart objects (e.g., in pie charts the observed increase in recall was more than 22\%, which, given the small sample number of pie charts in the test dataset,  corresponded to a reduction of the number of false negatives equal to 2). This may be explained by the fact that charts, especially pie charts, have more uniform shapes to start with, so CRFs can easily deduce the correct label given the saliency map and the input image. Line and bar charts yielded lower performance than pie charts, although the performance was still higher than the baseline. By comparing with the results obtained by SAL-CL model, we can infer that the lower performance was due to the difficulty by the saliency detector alone in extracting discriminative features between these two types of charts.

Indeed, this shortcoming was countered by introducing the classification loss in the model (SAL-CL configuration). The classifiers managed to aid the saliency detector network in recognizing the distinguishing features between line charts (increase to $F_1$ score of about 24\%) and bar charts (increase to $F_1$ score of about 23\%) w.r.t. the baseline, bringing the average $F_1$ score to 87\%.

\begin{figure*}[!ht]
	\begin{center}
		\includegraphics[width=7in]{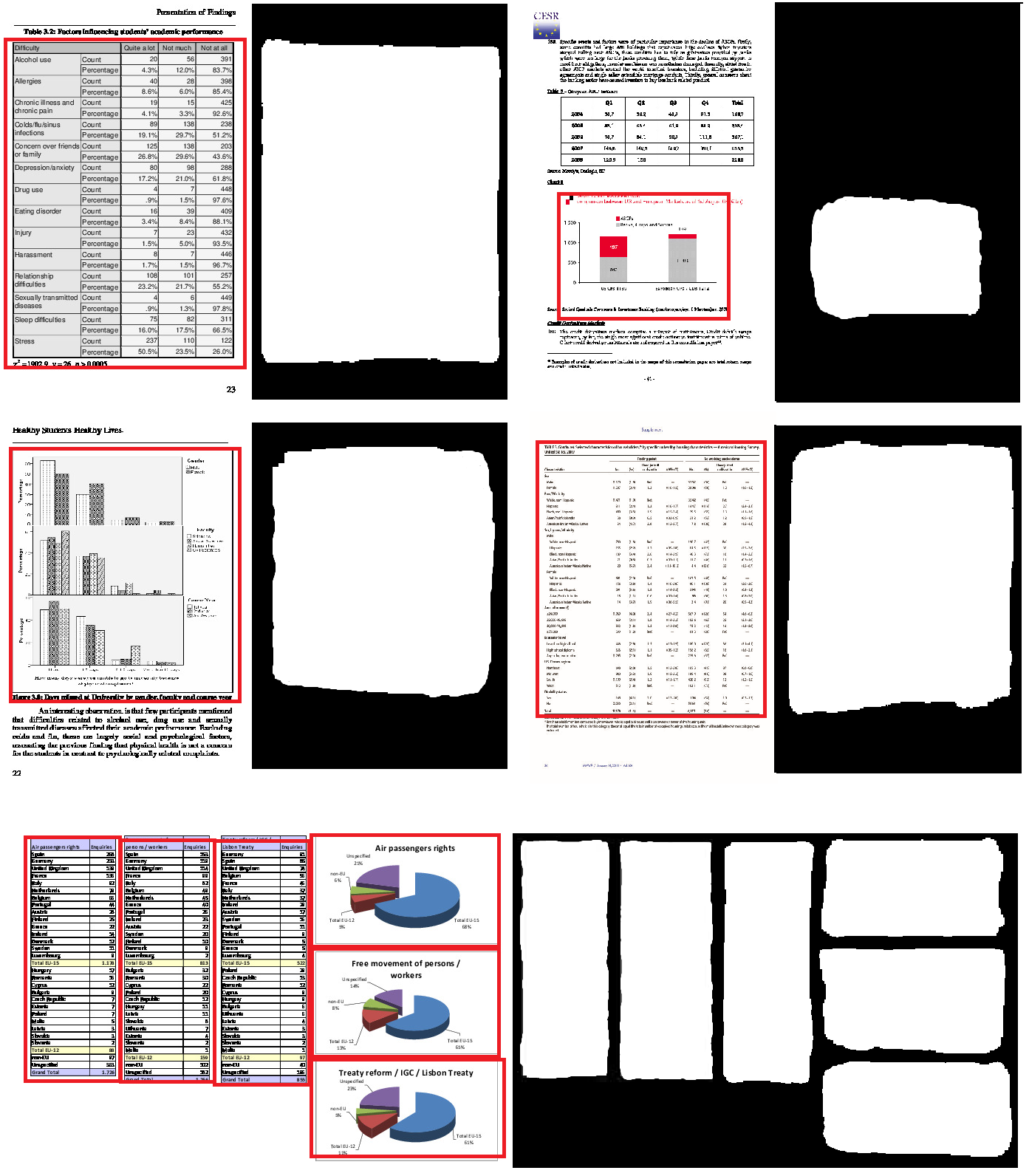}
		\caption{Examples of good detections by our method. \textit{Top row:} A ruled table and a bar chart; \textit{middle row:} a complex bar chart and an unruled table; \textit{Bottom row:} an image with three tables and three pie charts. The red bounding boxes superimposed on the original images are the final output of our system.}
		\label{example_good}
	\end{center}
\end{figure*}

\begin{figure*}[ht]
	\begin{center}
		\includegraphics[width=7in]{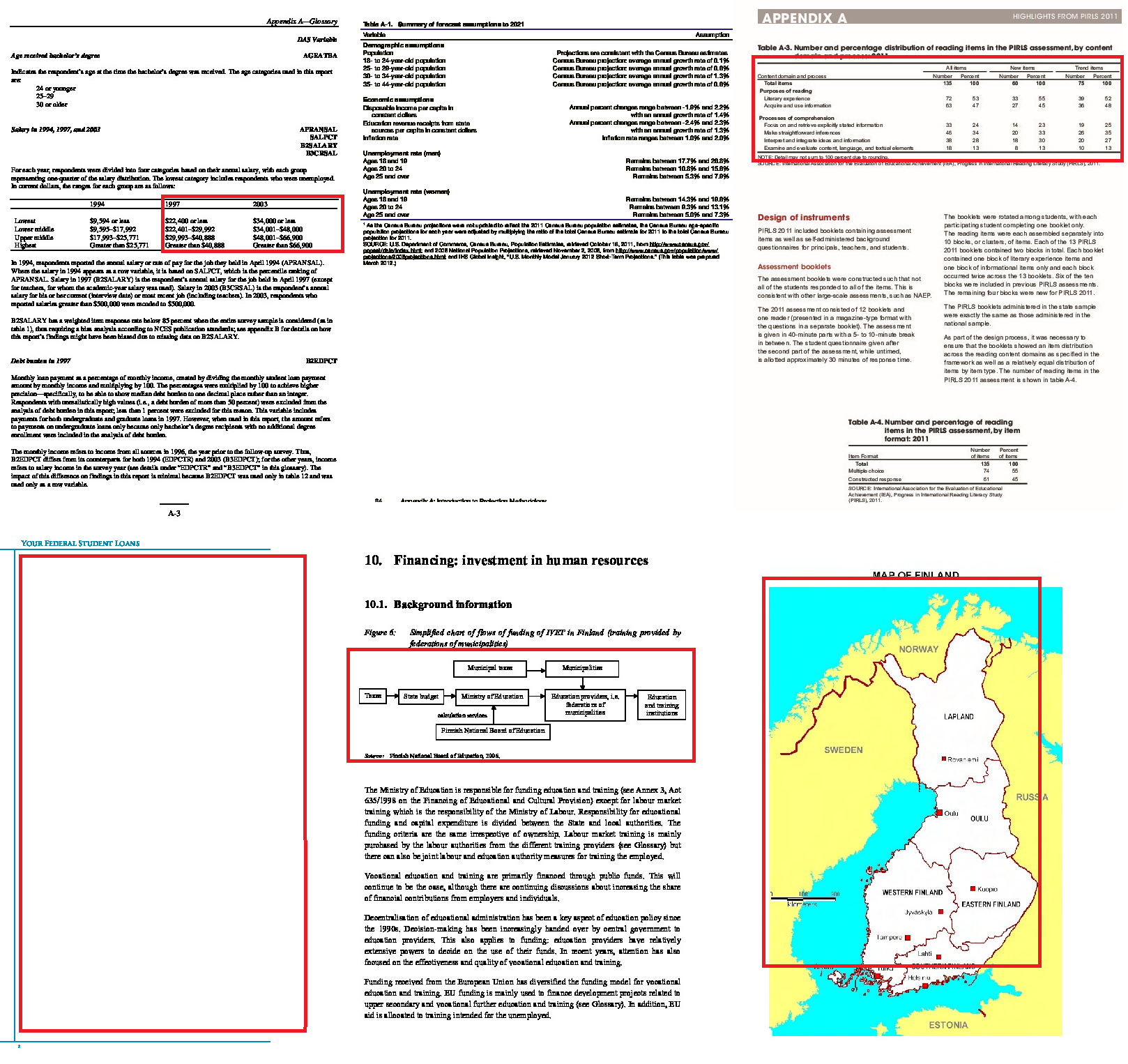}
		\caption{Examples of bad detections by our method. \textit{Top row, false negatives:}  A partially detected unruled table (left), a completely undetected unruled table (middle) and a page with two tables where one was not detected at all (the small one near the bottom). \textit{Bottom row, false positives:}  A complete white area that was recognized as a table (middle, the model was confused from the vertical and horizontal blue lines present in the same page), a flow diagram that was classified as a bar chart (middle) and an image that was classified as a line chart (left, partially).}
		\label{example_bad}
	\end{center}
\end{figure*}

Finally, testing the whole system (ALL) added a further 6.3\% increase to the system's performance, reaching a maximum average $F_1$ score of 93.4\%. Figures \ref{example_good} and \ref{example_bad} show examples of, respectively, good  and bad detections obtained by our method.

\begin{figure*}[ht]
	\begin{center}
		\includegraphics[width=7in]{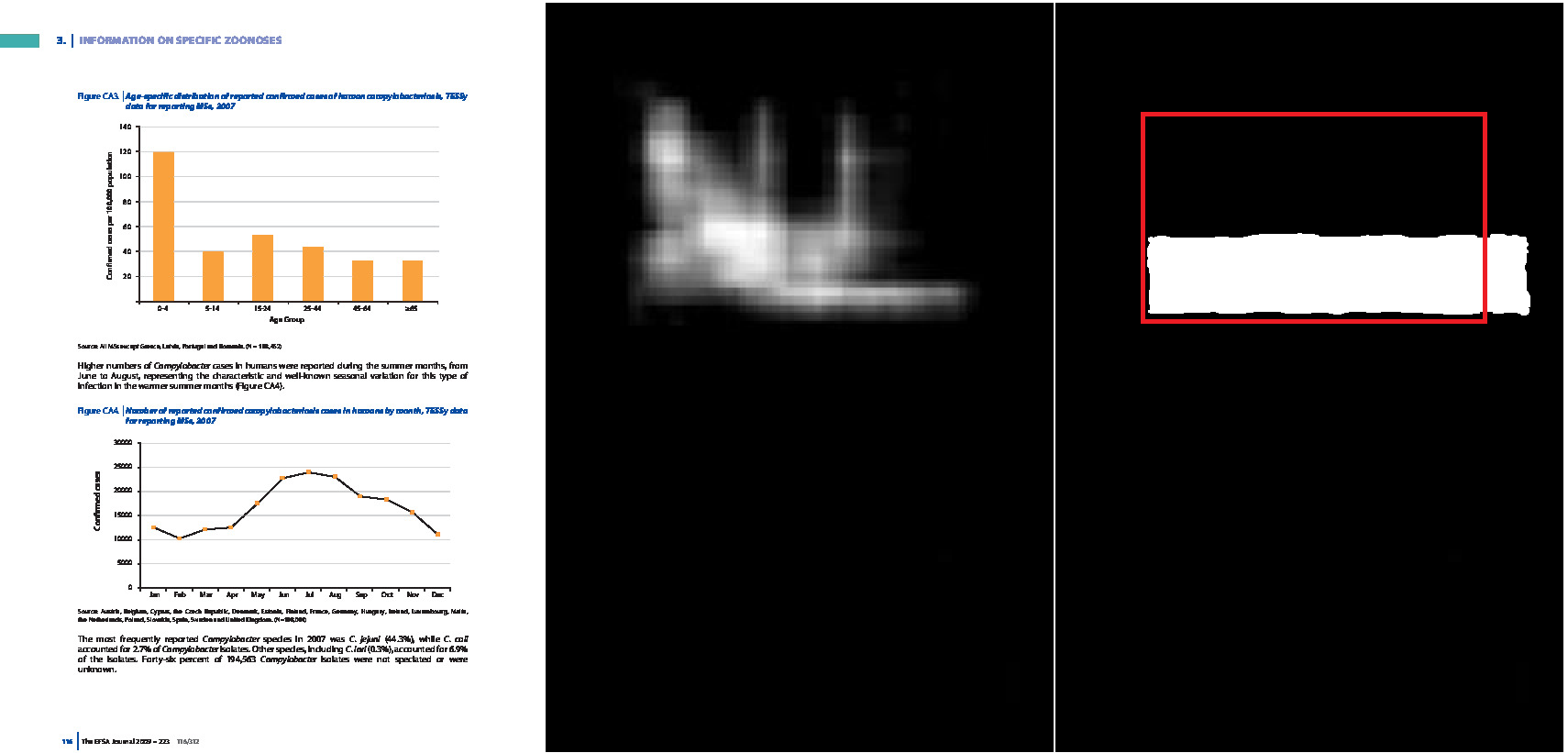}
		\caption{Partial detection of a bar chart by the network. The network correctly identified the most salient regions of the chart (middle) but failed to expand the detection's boundaries to cover it completely. The output shown in the middle, is the normalized result of the network before the CRF model was applied. The final output of the model is the white rectangle (right). The red bounding box represents the ground truth. }
		\label{example_partial}
	\end{center}
\end{figure*}

Overall, among the different chart types, pie chart detection was the one that mostly benefited from the CRF module, both because it is more easily distinguishable from bar charts and line charts, and because the relatively closed boundaries of pie charts are easily separable by the CRF's pairwise potential. For the same reason, the ``open'' structures of bar and line charts (as well as tables) make it harder to the CRF model to separate white areas belonging to the chart's region from white areas belonging to the background (e.g., Fig.~\ref{example_partial}). In these cases, the classification loss introduced with the Inception-based classifiers significantly helped in improving the corresponding accuracy scores.

\begin{table}[th]
	\begin{center}
		\caption{Performance --- in terms of Precision, Recall, F$_1$ and IOU --- achieved by the different configurations of our method on the extended ICDAR 2013 dataset. Values are in percentage.}
		\begin{tabular}{|c|c|c|c|c|c|}
			\hline
			Configuration & Class & Precision & Recall & $F_{1}$ &IOU\\			
			\hline\hline
			
			\multirow{4}{*}{SAL}
			&Tables &78.38&74.36&76.32&65.28\\ \cline{2-6}
			&Pie charts&75.00&66.67&70.59&62.67\\ \cline{2-6}
			&Bar charts&62.50&68.97&65.57&63.45\\ \cline{2-6}
			&Line charts&61.90&65.00&63.41&62.10\\ \cline{2-6}
			&Average &69.45&68.75&68.97&63.37\\
			
			\hline
			\hline
			
			\multirow{4}{*}{SAL--CRF}
			&Tables &81.88&83.97&82.91&71.39\\ \cline{2-6}
			&Pie charts&80.00&88.89&84.21&73.11\\ \cline{2-6}
			&Bar charts&73.33&75.86&74.58&68.55\\ \cline{2-6}
			&Line charts&68.00&73.91&70.83&67.87\\ \cline{2-6}
			&Average &75.80&80.66&78.13&70.23\\

			\hline
			\hline			
			\multirow{4}{*}{SAL--CL}
			&Tables &93.79&87.18&90.37&75.51\\ \cline{2-6}
			&Pie charts&87.50&77.78&82.35&72.22\\ \cline{2-6}
			&Bar charts&86.67&89.66&88.14&76.38\\ \cline{2-6}
			&Line charts&89.47&85.00&87.18&74.75\\ \cline{2-6}
			&Average &89.36&84.90&87.01&74.72\\
			
			\hline
			\hline
			\multirow{4}{*}{ALL}
			&Tables &97.45&98.08&97.76&81.33\\ \cline{2-6}
			&Pie charts&100&88.89&94.12&78.11\\ \cline{2-6}
			&Bar charts&90.00&93.10&91.53&79.59\\ \cline{2-6}
			&Line charts&90.00&90.00&90.00&78.50\\ \cline{2-6}
			&Average &94.36&92.52&93.35&79.38\\
			
			\hline
		\end{tabular}		
		\label{tab_performance_aal}
	\end{center}
\end{table}

To ground our work with state of the art in table detection, we compared the performance of the four different configurations of our method to those achieved by DeepDeSRT \cite{schreiber2017deepdesrt}, Tran \cite{tran2015table} and Hao \cite{hao2016table} in detecting only tables on the ICDAR 2013 dataset. The comparison is reported in Table~\ref{tab_performance_tables} and highlights the importance of the different components of our method. In the SAL configuration, the performance achieved were the lowest ($F_1$ = 76.3\%), meaning that generic saliency detection alone was not enough to completely extract all the information necessary to identify tables. There was a mild increase in performance when the CRF module was added (SAL-CRF configuration), where an increase of more than 6\% in $F_1$ score was observed ($F_1$ = 82.9\%). The increase was even more substantial when the classifiers were included in the configuration (SAL-CL), where an increase of 14\% in $F_1$ score was observed (90.4\%). When all components were included in the experiment (ALL configuration), our system achieved an $F_1$ score of 97.8\%, outperforming the state-of-the-art methods.

\begin{table}[th]
	\begin{center}
		\caption{Comparison of state of the art methods in table detection accuracy on the standard ICDAR 2013 dataset}
		\begin{tabular}{|c|c|c|c|}
			\hline
			& \multicolumn{3}{|c|}{Table Detection} \\
			\cline{2-4}
			& Precision & Recall & $F_{1}$ \\			
			\hline\hline
			DeepDeSRT \cite{schreiber2017deepdesrt} & 97.4 & 96.1 & 96.7\\
			\hline
			Tran \cite{tran2015table} & 95.2 & 96.4 & 95.8\\
			\hline
			Hao \cite{hao2016table} & 97.2 & 92.2 & 94.6\\
			\hline
			\hline
			SAL & 78.4 & 74.4 & 76.3\\
			\hline
			SAL-CRF & 81.9 & 83.4 & 82.9\\
			\hline
			SAL-CL & 93.8 & 87.2 & 90.4\\
			\hline 
			Our method (ALL) & 97.5 & 98.1 & 97.8 \\
			\hline
		\end{tabular}
		\label{tab_performance_tables}
	\end{center}
\end{table}                                                                                                                                                                                
What was clear from the results was that CRF-based approaches influence the number of false negatives and subsequently, the recall, more than they influence the number of false positives. In fact, w.r.t. the baseline configuration, the CRF module added a modest 6\% in precision, but 12\% in recall. This was explained by the fact that the major contribution of CRF models consisted in filling gaps and holes resulting from the deep learning methods, especially for very large tables with large white areas.

\section{Conclusion}
The identification of graphical elements such as tables and charts in documents is an essential processing block for any system that aims at extracting information automatically. In this paper, we presented a method for automatic table and chart detection in document files converted to images, hence without exploiting format information (e.g., PDF tokens or HTML tags) that limit the general applicability of these approaches. The core of our model is a DCNN trained to detect salient regions from document images, with saliency based on the categories of objects that we aim to identify (tables, pie charts, bar charts, line charts). An additional loss signal based on the generated saliency maps' discriminative power in a classification task was provided during training, and a fully-connected CRF model was finally employed to smooth and enhance the final outputs. Performance evaluation, carried out on the standard ICDAR 2013 benchmark and on an extended version with additional annotations of charts, showed that the proposed model achieves better performance than state-of-the-art methods in the localization of tables and charts.

\section*{Acknowledgment}

This work was funded by the Tab2Ex company, San Jose’, California (USA). We also gratefully acknowledge the support of NVIDIA Corporation for the donation of the Titan X Pascal GPUs used for this research.

\ifCLASSOPTIONcaptionsoff
  \newpage
\fi

\bibliographystyle{IEEEtran}
\bibliography{refs}

\begin{thebibliography}{10}
\providecommand{\url}[1]{#1}
\csname url@samestyle\endcsname
\providecommand{\newblock}{\relax}
\providecommand{\bibinfo}[2]{#2}
\providecommand{\BIBentrySTDinterwordspacing}{\spaceskip=0pt\relax}
\providecommand{\BIBentryALTinterwordstretchfactor}{4}
\providecommand{\BIBentryALTinterwordspacing}{\spaceskip=\fontdimen2\font plus
\BIBentryALTinterwordstretchfactor\fontdimen3\font minus
  \fontdimen4\font\relax}
\providecommand{\BIBforeignlanguage}[2]{{%
\expandafter\ifx\csname l@#1\endcsname\relax
\typeout{** WARNING: IEEEtran.bst: No hyphenation pattern has been}%
\typeout{** loaded for the language `#1'. Using the pattern for}%
\typeout{** the default language instead.}%
\else
\language=\csname l@#1\endcsname
\fi
#2}}
\providecommand{\BIBdecl}{\relax}
\BIBdecl

\bibitem{Redmon_2017_CVPR}
J.~Redmon and A.~Farhadi, ``Yolo9000: Better, faster, stronger,'' in \emph{The
  IEEE Conference on Computer Vision and Pattern Recognition (CVPR)}, July
  2017.

\bibitem{7792742}
J.~Li, Y.~Wei, X.~Liang, J.~Dong, T.~Xu, J.~Feng, and S.~Yan, ``Attentive
  contexts for object detection,'' \emph{IEEE Transactions on Multimedia},
  vol.~19, no.~5, pp. 944--954, May 2017.

\bibitem{ren2015faster}
S.~Ren, K.~He, R.~Girshick, and J.~Sun, ``Faster r-cnn: Towards real-time
  object detection with region proposal networks,'' in \emph{Advances in neural
  information processing systems}, 2015, pp. 91--99.

\bibitem{Girshick_2015_ICCV}
R.~Girshick, ``Fast r-cnn,'' in \emph{The IEEE International Conference on
  Computer Vision (ICCV)}, December 2015.

\bibitem{Itti1998}
L.~Itti, C.~Koch, and E.~Niebur, ``A model of saliency-based visual attention
  for rapid scene analysis,'' \emph{IEEE Transactions on Pattern Analysis and
  Machine Intelligence}, vol.~20, no.~11, pp. 1254--1259, Nov 1998.

\bibitem{YuKoltun2016}
F.~Yu and V.~Koltun, ``Multi-scale context aggregation by dilated
  convolutions,'' in \emph{International Conference on Learning
  Representations}, 2016.

\bibitem{MURABITO2018}
\BIBentryALTinterwordspacing
F.~Murabito, C.~Spampinato, S.~Palazzo, D.~Giordano, K.~Pogorelov, and
  M.~Riegler, ``Top-down saliency detection driven by visual classification,''
  \emph{Computer Vision and Image Understanding}, 2018. [Online]. Available:
  \url{http://www.sciencedirect.com/science/article/pii/S1077314218300407}
\BIBentrySTDinterwordspacing

\bibitem{NIPS2011_4296}
\BIBentryALTinterwordspacing
P.~Kr\"{a}henb\"{u}hl and V.~Koltun, ``Efficient inference in fully connected
  crfs with gaussian edge potentials,'' in \emph{Advances in Neural Information
  Processing Systems 24}, J.~Shawe-Taylor, R.~S. Zemel, P.~L. Bartlett,
  F.~Pereira, and K.~Q. Weinberger, Eds.\hskip 1em plus 0.5em minus 0.4em\relax
  Curran Associates, Inc., 2011, pp. 109--117. [Online]. Available:
  \url{http://papers.nips.cc/paper/4296-efficient-inference-in-fully-connected-crfs-with-gaussian-edge-potentials.pdf}
\BIBentrySTDinterwordspacing

\bibitem{chen2011table}
J.~Chen and D.~Lopresti, ``Table detection in noisy off-line handwritten
  documents,'' in \emph{Document Analysis and Recognition (ICDAR), 2011
  International Conference on}.\hskip 1em plus 0.5em minus 0.4em\relax IEEE,
  2011, pp. 399--403.

\bibitem{fang2011table}
J.~Fang, L.~Gao, K.~Bai, R.~Qiu, X.~Tao, and Z.~Tang, ``A table detection
  method for multipage pdf documents via visual seperators and tabular
  structures,'' in \emph{Document Analysis and Recognition (ICDAR), 2011
  International Conference on}.\hskip 1em plus 0.5em minus 0.4em\relax IEEE,
  2011, pp. 779--783.

\bibitem{shafait2010table}
F.~Shafait and R.~Smith, ``Table detection in heterogeneous documents,'' in
  \emph{Proceedings of the 9th IAPR International Workshop on Document Analysis
  Systems}.\hskip 1em plus 0.5em minus 0.4em\relax ACM, 2010, pp. 65--72.

\bibitem{gatos2005automatic}
B.~Gatos, D.~Danatsas, I.~Pratikakis, and S.~J. Perantonis, ``Automatic table
  detection in document images,'' in \emph{International Conference on Pattern
  Recognition and Image Analysis}.\hskip 1em plus 0.5em minus 0.4em\relax
  Springer, 2005, pp. 609--618.

\bibitem{deivalakshmi2014detection}
S.~Deivalakshmi, K.~Chaitanya, and P.~Palanisamy, ``Detection of table
  structure and content extraction from scanned documents,'' in
  \emph{Communications and Signal Processing (ICCSP), 2014 International
  Conference on}.\hskip 1em plus 0.5em minus 0.4em\relax IEEE, 2014, pp.
  270--274.

\bibitem{krupl2005using}
B.~Kr{\"u}pl, M.~Herzog, and W.~Gatterbauer, ``Using visual cues for extraction
  of tabular data from arbitrary html documents,'' in \emph{Special interest
  tracks and posters of the 14th international conference on World Wide
  Web}.\hskip 1em plus 0.5em minus 0.4em\relax ACM, 2005, pp. 1000--1001.

\bibitem{krupl2006visually}
B.~Kr{\"u}pl and M.~Herzog, ``Visually guided bottom-up table detection and
  segmentation in web documents,'' in \emph{Proceedings of the 15th
  international conference on World Wide Web}.\hskip 1em plus 0.5em minus
  0.4em\relax ACM, 2006, pp. 933--934.

\bibitem{jahan2014locating}
M.~A. Jahan and R.~G. Ragel, ``Locating tables in scanned documents for
  reconstructing and republishing,'' in \emph{Information and Automation for
  Sustainability (ICIAfS), 2014 7th International Conference on}.\hskip 1em
  plus 0.5em minus 0.4em\relax IEEE, 2014, pp. 1--6.

\bibitem{liu2008identifying}
Y.~Liu, P.~Mitra, and C.~L. Giles, ``Identifying table boundaries in digital
  documents via sparse line detection,'' in \emph{Proceedings of the 17th ACM
  conference on Information and knowledge management}.\hskip 1em plus 0.5em
  minus 0.4em\relax ACM, 2008, pp. 1311--1320.

\bibitem{raskovic2012borderless}
M.~Raskovic, N.~Bozidarevic, and M.~Sesum, ``Borderless table detection
  engine,'' Jan.~23 2012, uS Patent App. 13/521,424.

\bibitem{huang2003model}
W.~Huang, C.~L. Tan, and W.~K. Leow, ``Model-based chart image recognition,''
  in \emph{International Workshop on Graphics Recognition}.\hskip 1em plus
  0.5em minus 0.4em\relax Springer, 2003, pp. 87--99.

\bibitem{shao2005recognition}
M.~Shao and R.~P. Futrelle, ``Recognition and classification of figures in pdf
  documents,'' in \emph{International Workshop on Graphics Recognition}.\hskip
  1em plus 0.5em minus 0.4em\relax Springer, 2005, pp. 231--242.

\bibitem{huang2007extraction}
W.~Huang, R.~Liu, and C.-L. Tan, ``Extraction of vectorized graphical
  information from scientific chart images,'' in \emph{Document Analysis and
  Recognition, 2007. ICDAR 2007. Ninth International Conference on},
  vol.~1.\hskip 1em plus 0.5em minus 0.4em\relax IEEE, 2007, pp. 521--525.

\bibitem{karthikeyani2012machine}
V.~Karthikeyani and S.~Nagarajan, ``Machine learning classification algorithms
  to recognize chart types in portable document format (pdf) files,''
  \emph{International Journal of Computer Applications}, vol.~39, no.~2, 2012.

\bibitem{perez2016tao}
M.~O. Perez-Arriaga, T.~Estrada, and S.~Abad-Mota, ``Tao: System for table
  detection and extraction from pdf documents.'' in \emph{FLAIRS Conference},
  2016, pp. 591--596.

\bibitem{shetty2007segmentation}
S.~Shetty, H.~Srinivasan, M.~Beal, and S.~Srihari, ``Segmentation and labeling
  of documents using conditional random fields,'' in \emph{Document Recognition
  and Retrieval XIV}, vol. 6500.\hskip 1em plus 0.5em minus 0.4em\relax
  International Society for Optics and Photonics, 2007, p. 65000U.

\bibitem{delaye2012text}
A.~Delaye and C.-L. Liu, ``Text/non-text classification in online handwritten
  documents with conditional random fields,'' in \emph{Chinese Conference on
  Pattern Recognition}.\hskip 1em plus 0.5em minus 0.4em\relax Springer, 2012,
  pp. 514--521.

\bibitem{pinto2003table}
D.~Pinto, A.~McCallum, X.~Wei, and W.~B. Croft, ``Table extraction using
  conditional random fields,'' in \emph{Proceedings of the 26th annual
  international ACM SIGIR conference on Research and development in informaion
  retrieval}.\hskip 1em plus 0.5em minus 0.4em\relax ACM, 2003, pp. 235--242.

\bibitem{hao2016table}
L.~Hao, L.~Gao, X.~Yi, and Z.~Tang, ``A table detection method for pdf
  documents based on convolutional neural networks,'' in \emph{Document
  Analysis Systems (DAS), 2016 12th IAPR Workshop on}.\hskip 1em plus 0.5em
  minus 0.4em\relax IEEE, 2016, pp. 287--292.

\bibitem{afzal2015deepdocclassifier}
M.~Z. Afzal, S.~Capobianco, M.~I. Malik, S.~Marinai, T.~M. Breuel, A.~Dengel,
  and M.~Liwicki, ``Deepdocclassifier: Document classification with deep
  convolutional neural network,'' in \emph{Document Analysis and Recognition
  (ICDAR), 2015 13th International Conference on}.\hskip 1em plus 0.5em minus
  0.4em\relax IEEE, 2015, pp. 1111--1115.

\bibitem{kang2014convolutional}
L.~Kang, J.~Kumar, P.~Ye, Y.~Li, and D.~Doermann, ``Convolutional neural
  networks for document image classification,'' in \emph{Pattern Recognition
  (ICPR), 2014 22nd International Conference on}.\hskip 1em plus 0.5em minus
  0.4em\relax IEEE, 2014, pp. 3168--3172.

\bibitem{roy2016generalized}
S.~Roy, A.~Das, and U.~Bhattacharya, ``Generalized stacking of
  layerwise-trained deep convolutional neural networks for document image
  classification,'' in \emph{Pattern Recognition (ICPR), 2016 23rd
  International Conference on}.\hskip 1em plus 0.5em minus 0.4em\relax IEEE,
  2016, pp. 1273--1278.

\bibitem{liu2015chart}
X.~Liu, B.~Tang, Z.~Wang, X.~Xu, S.~Pu, D.~Tao, and M.~Song, ``Chart
  classification by combining deep convolutional networks and deep belief
  networks,'' in \emph{Document Analysis and Recognition (ICDAR), 2015 13th
  International Conference on}.\hskip 1em plus 0.5em minus 0.4em\relax IEEE,
  2015, pp. 801--805.

\bibitem{tang2016deepchart}
B.~Tang, X.~Liu, J.~Lei, M.~Song, D.~Tao, S.~Sun, and F.~Dong, ``Deepchart:
  Combining deep convolutional networks and deep belief networks in chart
  classification,'' \emph{Signal Processing}, vol. 124, pp. 156--161, 2016.

\bibitem{gilani2017table}
A.~Gilani, S.~R. Qasim, and F.~Malik, Imran abd~Shafait, ``Table detection
  using deep learning,'' in \emph{Document Analysis and Recognition (ICDAR),
  2017 16th International Conference on}.\hskip 1em plus 0.5em minus
  0.4em\relax IEEE, 2017.

\bibitem{tran2015table}
D.~N. Tran, T.~A. Tran, A.~Oh, S.~H. Kim, and I.~S. Na, ``Table detection from
  document image using vertical arrangement of text blocks,''
  \emph{International Journal of Contents}, vol.~11, no.~4, pp. 77--85, 2015.

\bibitem{li2016visual}
G.~Li and Y.~Yu, ``Visual saliency detection based on multiscale deep cnn
  features,'' \emph{IEEE Transactions on Image Processing}, vol.~25, no.~11,
  pp. 5012--5024, 2016.

\bibitem{kummerer2014deep}
M.~K{\"u}mmerer, L.~Theis, and M.~Bethge, ``Deep gaze i: Boosting saliency
  prediction with feature maps trained on imagenet,'' \emph{arXiv preprint
  arXiv:1411.1045}, 2014.

\bibitem{jiang2015salicon}
M.~Jiang, S.~Huang, J.~Duan, and Q.~Zhao, ``Salicon: Saliency in context,'' in
  \emph{Computer Vision and Pattern Recognition (CVPR), 2015 IEEE Conference
  on}.\hskip 1em plus 0.5em minus 0.4em\relax IEEE, 2015, pp. 1072--1080.

\bibitem{he2015supercnn}
S.~He, R.~W. Lau, W.~Liu, Z.~Huang, and Q.~Yang, ``Supercnn: A superpixelwise
  convolutional neural network for salient object detection,''
  \emph{International journal of computer vision}, vol. 115, no.~3, pp.
  330--344, 2015.

\bibitem{liu2016dhsnet}
N.~Liu and J.~Han, ``Dhsnet: Deep hierarchical saliency network for salient
  object detection,'' in \emph{Computer Vision and Pattern Recognition (CVPR),
  2016 IEEE Conference on}.\hskip 1em plus 0.5em minus 0.4em\relax IEEE, 2016,
  pp. 678--686.

\bibitem{li2015efficient}
Y.~Li, K.~Fu, Z.~Liu, and J.~Yang, ``Efficient saliency-model-guided visual
  co-saliency detection,'' \emph{IEEE Signal Processing Letters}, vol.~22,
  no.~5, pp. 588--592, 2015.

\bibitem{zhao2015saliency}
R.~Zhao, W.~Ouyang, H.~Li, and X.~Wang, ``Saliency detection by multi-context
  deep learning,'' in \emph{Proceedings of the IEEE Conference on Computer
  Vision and Pattern Recognition}, 2015, pp. 1265--1274.

\bibitem{pan2016shallow}
J.~Pan, E.~Sayrol, X.~Giro-i Nieto, K.~McGuinness, and N.~E. O'Connor,
  ``Shallow and deep convolutional networks for saliency prediction,'' in
  \emph{Proceedings of the IEEE Conference on Computer Vision and Pattern
  Recognition}, 2016, pp. 598--606.

\bibitem{Almahairi2016}
\BIBentryALTinterwordspacing
A.~Almahairi, N.~Ballas, T.~Cooijmans, Y.~Zheng, H.~Larochelle, and
  A.~Courville, ``Dynamic capacity networks,'' in \emph{Proceedings of the 33rd
  International Conference on International Conference on Machine Learning -
  Volume 48}, ser. ICML'16.\hskip 1em plus 0.5em minus 0.4em\relax JMLR.org,
  2016, pp. 2549--2558. [Online]. Available:
  \url{http://dl.acm.org/citation.cfm?id=3045390.3045659}
\BIBentrySTDinterwordspacing

\bibitem{6247739}
J.~Yao, S.~Fidler, and R.~Urtasun, ``Describing the scene as a whole: Joint
  object detection, scene classification and semantic segmentation,'' in
  \emph{2012 IEEE Conference on Computer Vision and Pattern Recognition}, June
  2012, pp. 702--709.

\bibitem{long2015fully}
J.~Long, E.~Shelhamer, and T.~Darrell, ``Fully convolutional networks for
  semantic segmentation,'' in \emph{Proceedings of the IEEE conference on
  computer vision and pattern recognition}, 2015, pp. 3431--3440.

\bibitem{Simonyan14c}
K.~Simonyan and A.~Zisserman, ``Very deep convolutional networks for
  large-scale image recognition,'' \emph{ICLR 2015}, vol. abs/1409.1556, 2014.

\bibitem{szegedy2015going}
C.~Szegedy, W.~Liu, Y.~Jia, P.~Sermanet, S.~Reed, D.~Anguelov, D.~Erhan,
  V.~Vanhoucke, A.~Rabinovich \emph{et~al.}, ``Going deeper with
  convolutions.''\hskip 1em plus 0.5em minus 0.4em\relax Cvpr, 2015.

\bibitem{kohli2009robust}
P.~Kohli, P.~H. Torr \emph{et~al.}, ``Robust higher order potentials for
  enforcing label consistency,'' \emph{International Journal of Computer
  Vision}, vol.~82, no.~3, pp. 302--324, 2009.

\bibitem{chen2016deeplab}
L.-C. Chen, G.~Papandreou, I.~Kokkinos, K.~Murphy, and A.~L. Yuille, ``Deeplab:
  Semantic image segmentation with deep convolutional nets, atrous convolution,
  and fully connected crfs,'' \emph{arXiv preprint arXiv:1606.00915}, 2016.

\bibitem{liu2015crf}
F.~Liu, G.~Lin, and C.~Shen, ``Crf learning with cnn features for image
  segmentation,'' \emph{Pattern Recognition}, vol.~48, no.~10, pp. 2983--2992,
  2015.

\bibitem{krahenbuhl2011efficient}
P.~Kr{\"a}henb{\"u}hl and V.~Koltun, ``Efficient inference in fully connected
  crfs with gaussian edge potentials,'' in \emph{Advances in neural information
  processing systems}, 2011, pp. 109--117.

\bibitem{Chen2015}
L.~Chen, G.~Papandreou, I.~Kokkinas, K.~Murphy, and A.~Yuille, ``Semantic image
  segmentation with deep convolutional nets and fully connected crfs,''
  \emph{ICLR 2015}, vol. abs/1412.7062, 2015.

\bibitem{wu1982potts}
F.-Y. Wu, ``The potts model,'' \emph{Reviews of modern physics}, vol.~54,
  no.~1, p. 235, 1982.

\bibitem{karatzas2013icdar}
D.~Karatzas, F.~Shafait, S.~Uchida, M.~Iwamura, L.~G. i~Bigorda, S.~R. Mestre,
  J.~Mas, D.~F. Mota, J.~A. Almazan, and L.~P. de~las Heras, ``Icdar 2013
  robust reading competition,'' in \emph{Document Analysis and Recognition
  (ICDAR), 2013 12th International Conference on}.\hskip 1em plus 0.5em minus
  0.4em\relax IEEE, 2013, pp. 1484--1493.

\bibitem{kavasidis2014innovative}
I.~Kavasidis, S.~Palazzo, R.~Di~Salvo, D.~Giordano, and C.~Spampinato, ``An
  innovative web-based collaborative platform for video annotation,''
  \emph{Multimedia Tools and Applications}, vol.~70, no.~1, pp. 413--432, 2014.

\bibitem{deng2009imagenet}
J.~Deng, W.~Dong, R.~Socher, L.-J. Li, K.~Li, and L.~Fei-Fei, ``Imagenet: A
  large-scale hierarchical image database,'' in \emph{Computer Vision and
  Pattern Recognition, 2009. CVPR 2009. IEEE Conference on}.\hskip 1em plus
  0.5em minus 0.4em\relax IEEE, 2009, pp. 248--255.

\bibitem{kingma2014adam}
D.~Kingma and J.~Ba, ``Adam: A method for stochastic optimization,''
  \emph{arXiv preprint arXiv:1412.6980}, 2014.

\bibitem{everingham2010pascal}
M.~Everingham, L.~Van~Gool, C.~K. Williams, J.~Winn, and A.~Zisserman, ``The
  pascal visual object classes (voc) challenge,'' \emph{International journal
  of computer vision}, vol.~88, p. 303{\textendash}338, 2010.

\bibitem{schreiber2017deepdesrt}
S.~Schreiber, S.~Agne, I.~Wolf, A.~Dengel, and S.~Ahmed, ``Deepdesrt: Deep
  learning for detection and structure recognition of tables in document
  images,'' in \emph{Document Analysis and Recognition (ICDAR), 2017 14th IAPR
  International Conference on}, vol.~1.\hskip 1em plus 0.5em minus 0.4em\relax
  IEEE, 2017, pp. 1162--1167.

\end{thebibliography}

%
%
%
%
%
%



\end{document}